%% file: example_paper.tex
\documentclass[runningheads]{llncs}

\usepackage[T1]{fontenc}
\usepackage{graphicx}
\usepackage{subcaption}
\usepackage{booktabs}
\usepackage{hyperref}
\usepackage{amsmath}
\usepackage{amssymb}
\usepackage{mathtools}
\usepackage{wrapfig}
\usepackage[capitalize,noabbrev]{cleveref}

\usepackage[textsize=tiny]{todonotes}

\begin{document}

\title{MedQ-UNI: Toward Unified Medical Image Quality Assessment and Restoration via Vision-Language Modeling}

\titlerunning{MedQ-UNI}

\author{Jiyao Liu\inst{1}$^*$ \and
Junzhi Ning \inst{2}$^*$  \and 
Wanying Qu \inst{1}  \and 
Lihao Liu\inst{2}  \and 
Chenglong Ma\inst{1}  \and  \\
Junjun He \inst{2}$^\dagger$  \and 
Ningsheng Xu\inst{1}$^\dagger$
}

\authorrunning{J Liu et al. }

\institute{Fudan University, Shanghai, China \and
Shanghai Artificial Intelligence Laboratory, Shanghai, China}

\maketitle

\renewcommand\thefootnote{}
\footnotetext{$^*$Equal contribution. $^\dagger$Corresponding author. }
\renewcommand\thefootnote{\arabic{footnote}}

\input{sections/abstract}

\input{sections/introduction}

\input{sections/task_paradigm}

\input{sections/model_design}

\input{sections/experiments}

\input{sections/conclusion}

\clearpage
\newpage

\bibliographystyle{splncs04}
\bibliography{example_paper}

\end{document}

%% file: sections/abstract.tex
\begin{abstract}
Existing medical image restoration (Med-IR) methods are typically modality-specific or degradation-specific, failing to generalize across the heterogeneous degradations encountered in clinical practice. We argue this limitation stems from the isolation of Med-IR from medical image quality assessment (Med-IQA), as restoration models without explicit quality understanding struggle to adapt to diverse degradation types across modalities. To address these challenges, we propose \textbf{MedQ-UNI}, a unified vision-language model that follows an \emph{assess-then-restore} paradigm, explicitly leveraging Med-IQA to guide Med-IR across arbitrary modalities and degradation types. MedQ-UNI adopts a multimodal autoregressive dual-expert architecture with shared attention: a quality assessment expert first identifies degradation issues through structured natural language descriptions, and a restoration expert then conditions on these descriptions to perform targeted image restoration. To support this paradigm, we construct a large-scale dataset of approximately 50K paired samples spanning three imaging modalities and five restoration tasks, each annotated with structured quality descriptions for joint Med-IQA and Med-IR training, along with a 2K-sample benchmark for evaluation. Extensive experiments demonstrate that a single MedQ-UNI model, without any task-specific adaptation, achieves state-of-the-art restoration performance across all tasks while generating superior descriptions, confirming that explicit quality understanding meaningfully improves restoration fidelity and interpretability.
\end{abstract}

%% file: sections/introduction.tex
\section{Introduction}

\begin{figure}[t]
\centering
\includegraphics[width=0.9\textwidth]{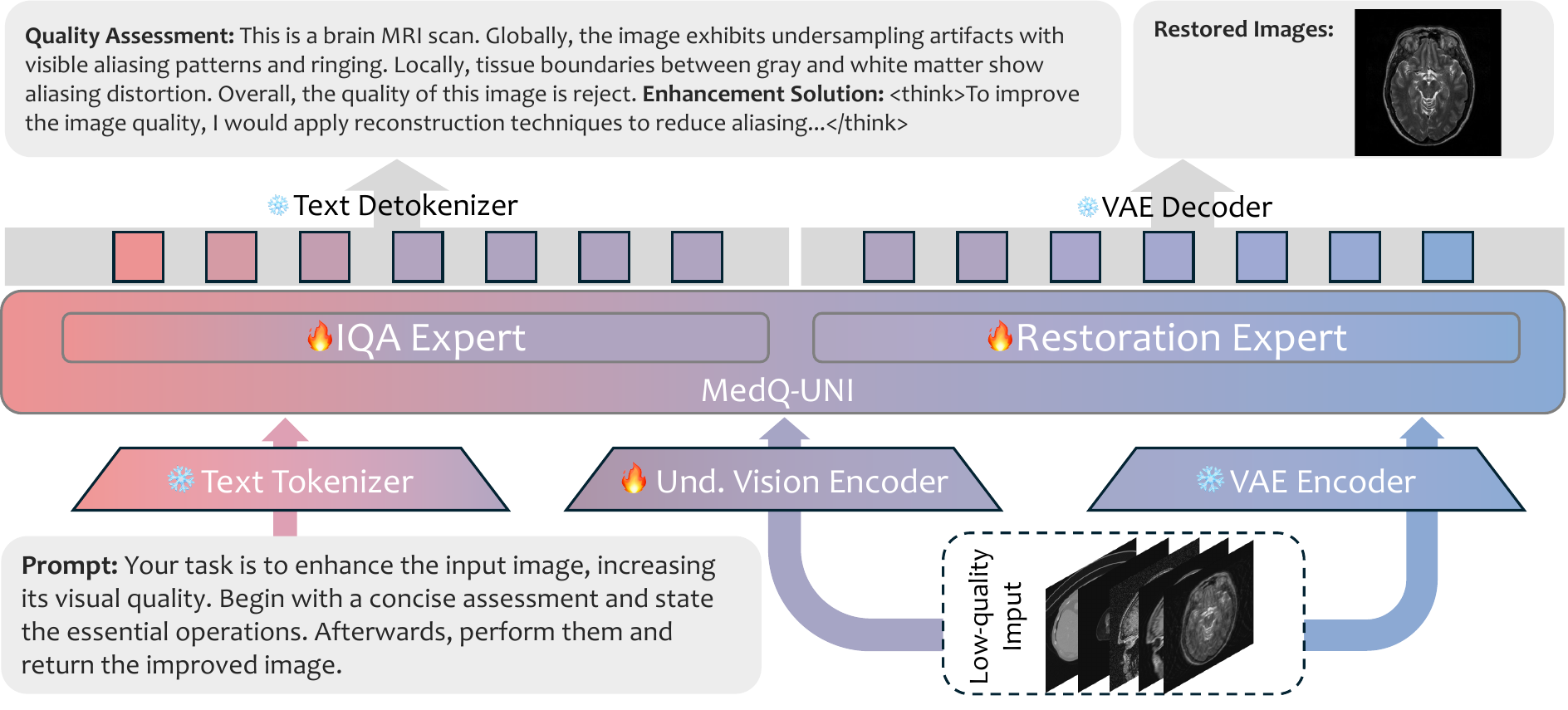}
\caption{Overview of MedQ-UNI. Given a degraded medical image and a task instruction, the Med-IQA expert first produces a structured natural language quality description, which then explicitly conditions the Med-IR expert via shared attention to generate the restored image. Both experts reside within a single multimodal autoregressive model.}
\label{fig:architecture}
\vspace{-1em}
\end{figure}

Medical images acquired across modalities such as CT, MRI, and PET are routinely degraded by heterogeneous degradations due to hardware limitations and acquisition constraints, directly compromising downstream analysis~\cite{ahishakiye2021survey}. Yet existing medical image restoration (Med-IR) methods~\cite{medical_denoising_survey2024,chungdecomposed} are typically modality-specific or degradation-specific, failing to generalize across the diverse imaging conditions. A central reason is the longstanding isolation of Med-IR from medical image quality assessment (Med-IQA)~\cite{pns_miqa2024,nr_iqa_ct2024,iqagpt2024}: Med-IQA methods predominantly produce scalar scores or discrete labels that lack descriptive degradation characterizations, while Med-IR methods operate ``blindly'' without awareness of actual image quality or modality-specific degradation characteristics. This disconnect mismatches the clinical workflow, where quality assessment and restoration are inherently coupled, and calls for a unified framework that first understands degradation and then leverages that understanding to drive targeted, generalizable restoration.

Recent work in the natural image domain provides a promising direction. Vision-language models such as DepictQA~\cite{depictqa2024} can describe quality issues in natural language rather than abstract scores, while DA-CLIP~\cite{daclip2024} and InstructIR~\cite{instructir2024} show that language-guided conditioning substantially improves restoration. Furthermore, unified understanding-generation architectures~\cite{wang2026multimodal,unifiedio2024} demonstrate a consistent finding: \emph{stronger understanding of visual content leads to better generation}. These results suggest that explicitly coupling quality understanding with restoration is not merely convenient but fundamentally beneficial. In the medical domain, recent all-in-one Med-IR methods such as AMIR~\cite{amir_medical2024} have begun to address multi-task restoration within a single model, yet they focus exclusively on pixel-level reconstruction. 
This calls for a unified model that couples quality understanding with generalizable restoration across modalities and degradation types.

To address this, we propose \textbf{MedQ-UNI}, a \emph{single} multimodal autoregressive model that realizes the \emph{assess-then-restore} paradigm for medical imaging. Within this unified model, a Med-IQA expert first autonomously perceives the input image and generates structured natural language quality descriptions that identify degradation type and severity, enabling the model to self-determine the appropriate restoration strategy without manual specification. A Med-IR expert then leverages these descriptions as conditioning signals to perform targeted restoration. Both experts reside in a dual-expert architecture with shared attention, forming an explicit causal chain from quality understanding to image generation that is interpretable and controllable across arbitrary modalities and degradation types.

Our main contributions are threefold: (1) we formalize the \emph{assess-then-restore} paradigm and propose MedQ-UNI, the first unified vision-language framework coupling Med-IQA with Med-IR across five tasks and three modalities through a single network; (2) we design a multimodal autoregressive dual-expert architecture with shared attention and a two-stage training strategy for interpretable, quality-conditioned restoration; and (3) we construct a 50K-sample multi-modal dataset with structured quality descriptions and a 2K-sample benchmark. Our results demonstrate that MedQ-UNI achieves state-of-the-art restoration, while its generated quality descriptions outperform those of GPT-4o. 

%% file: sections/task_paradigm.tex
\section{Task Paradigm and Dataset Construction}

\subsection{Task Paradigm}\label{ssec:task-paradigm}

MedQ-UNI operates under the \emph{assess-then-restore} paradigm, establishing an explicit causal chain: quality understanding $\rightarrow$ restoration strategy $\rightarrow$ image generation. Given a low-quality medical image $I_{\text{low}} \in \mathbb{R}^{H \times W \times C}$ and an optional text instruction $\mathcal{T}_{\text{inst}}$, the model first generates a structured natural language quality description $\mathcal{T}_q$ and then produces a restored image $I_{\text{high}}$:
\begin{align}
\mathcal{T}_q &= f_{\text{IQA}}(I_{\text{low}}; \theta_{\text{IQA}}), \\
I_{\text{high}} &= f_{\text{restore}}(I_{\text{low}}, \mathcal{T}_q; \theta_{\text{restore}}),
\end{align}
where the quality description $\mathcal{T}_q$ serves as an intermediate conditioning signal for restoration. The description $\mathcal{T}_q$ systematically covers five aspects: (i) modality and anatomical region identification; (ii) degradation characterization including type and severity; (iii) technical attribution of underlying causes; (iv) assessment of diagnostic impact; and (v) quality judgment with a Good/Usable/Reject recommendation. The unified formulation can be compactly expressed as:
\begin{equation}
(\mathcal{T}_q, I_{\text{high}}) = f_{\text{unified}}(I_{\text{low}}, \mathcal{T}_{\text{inst}}; \theta),
\end{equation}
where $\theta = \{\theta_{\text{IQA}}, \theta_{\text{restore}}, \theta_{\text{shared}}\}$ includes assessment-specific, restoration-specific, and shared parameters. The model jointly outputs both $\mathcal{T}_q$ and $I_{\text{high}}$, making the restoration process interpretable and controllable.

\subsection{Dataset Construction}\label{ssec:dataset-construction}

To enable joint Med-IQA and Med-IR training under the assess-then-restore paradigm, we construct a large-scale, multi-modal dataset that, to our knowledge, is the first to pair medical image restoration samples with structured natural language quality descriptions.

\noindent \textbf{Data sources and restoration pairs.} We curate paired low-quality / high-quality images from three public datasets spanning three modalities and five restoration tasks: (1) the 2016 NIH AAPM-Mayo Clinic Low-Dose CT Grand Challenge dataset~\cite{mccollough2017low} for low-dose CT denoising; (2) the IXI brain MRI dataset~\cite{ixi_dataset}, from which we derive three tasks: undersampled MRI reconstruction, motion artifact removal, and denoising; and (3) the UDPET ultra-low-dose PET challenge dataset~\cite{xue2025udpet} for PET denoising. 

\noindent \textbf{Quality description annotation.} For each low-quality image, we produce a structured quality description following the five-aspect protocol defined in Sec.~\ref{ssec:task-paradigm}. We adopt an iterative pseudo-label refinement pipeline: Qwen3-VL-235B~\cite{yang2025qwen3vl} first generates initial descriptions, clinical experts then review and correct them, and the refined annotations are fed back to improve subsequent generation rounds. All generated descriptions are then reviewed and corrected by three independent clinical experts per image, with final annotations reconciled through majority agreement following the annotation protocol of MedQbench~\cite{liu2025medq}. Every training and benchmark sample thus contains a degraded image, its high-quality reference, and a structured quality description, enabling the model to jointly learn quality understanding and image restoration.

%% file: sections/model_design.tex
\section{Model Design}

\subsection{Network Architecture}

MedQ-UNI is built upon a unified multimodal autoregressive architecture design~\cite{ning2025unimedvl}, as shown in Fig.~\ref{fig:architecture}. Specifically, it employs two full-capacity experts operating on a single interleaved token stream. The two experts share self-attention layers but maintain independent feed-forward networks, enabling quality understanding to explicitly guide restoration without task interference. The \textbf{Med-IQA expert} receives text tokens from a text tokenizer and image tokens from an understanding visual encoder, and autoregressively generates the quality description $\mathcal{T}_q$; the resulting quality tokens $\mathbf{Q}$ are retained in the shared attention context. The \textbf{Med-IR expert} receives VAE latent tokens $\mathbf{Z}_{\text{low}} = \text{VAE}_{\text{enc}}(I_{\text{low}})$ and produces the restored latent $\mathbf{Z}_{\text{high}}$, which is decoded as $I_{\text{high}} = \text{VAE}_{\text{dec}}(\mathbf{Z}_{\text{high}})$. Both experts share self-attention in every transformer block, so the Med-IR expert can attend to quality tokens $\mathbf{Q}$ produced by the Med-IQA expert. Independent feed-forward parameters per expert reduce gradient interference between the two tasks, effectively alleviating task conflict while preserving cross-modal alignment through the shared attention context.

\subsection{Two-Stage Training}

We adopt a two-stage training strategy to ensure stable optimization. Each training sample is organized as $\mathbf{x} = (\mathbf{x}_\text{con}, \mathbf{x}_\text{res})$, where $\mathbf{x}_\text{con}$ denotes the condition (task instruction and input image) and $\mathbf{x}_\text{res}$ denotes the response (restoration target or quality description). Let $l_\text{con}$ and $l_\text{res}$ be the condition and response lengths, respectively, and $\theta$ be all trainable parameters.

\noindent \textbf{Stage~1: Restoration pre-training.} In the first stage, only the Med-IR expert is trained to acquire strong pixel-level restoration capabilities before introducing quality understanding. The condition $\mathbf{x}_\text{con}$ contains the input instruction and low-quality image $I_{\text{low}}$, and the response $\mathbf{x}_\text{res}$ is the restored image $I_{\text{high}}$. The Med-IR expert is optimized with a rectified flow objective in the VAE latent space:
\begin{equation}
\mathcal{L}_{\text{restore}}(\theta) = \mathbb{E}_{\mathbf{x}\sim \mathcal{D}_\text{restore},\mathbf{z}_0\sim\mathcal{N}(0,\mathbf{I})}
 \left[\|v_{\theta}(\mathbf{z}_t, t | \mathbf{x}_\text{con})-(\mathbf{x}_\text{res}-\mathbf{z}_0)\|^2 \right],
\end{equation}
where $\mathbf{z}_t=t\mathbf{x}_\text{res}+(1-t) \mathbf{z}_0$, $v_\theta$ denotes the velocity network, $\mathcal{D}_\text{restore}$ denotes the restoration data, and $t$ is the sampled diffusion timestep. To further enforce pixel-level fidelity and structural preservation, we augment this with an L1 pixel loss and an SSIM loss computed in image space:
\begin{equation}
\mathcal{L}_{\text{pixel}}(\theta) = \|I_{\text{high}} - I_{\text{gt}}\|_1, \quad \mathcal{L}_{\text{SSIM}}(\theta) = 1 - \text{SSIM}(I_{\text{high}}, I_{\text{gt}}),
\end{equation}
where $I_{\text{gt}}$ is the ground-truth image. The full Stage~1 objective is:
\begin{equation}
\mathcal{L}_{\text{IR}}(\theta) = \mathcal{L}_{\text{restore}}(\theta) + \alpha \mathcal{L}_{\text{pixel}}(\theta) + \beta \mathcal{L}_{\text{SSIM}}(\theta),
\end{equation}
where $\alpha=0.2$ and $\beta=0.1$ balance the contributions of each term.

\noindent \textbf{Stage~2: Joint Med-IQA and Med-IR training.} In the second stage, the Med-IQA expert is activated and jointly trained alongside the Med-IR expert. For assessment-guided restoration, $\mathbf{x}_\text{con}$ additionally includes the generated quality description $\mathcal{T}_q$. The Med-IQA expert is trained with an autoregressive maximum likelihood objective over the quality description tokens:
\begin{equation}
\mathcal{L}_{\text{IQA}}(\theta) = -\mathbb{E}_{\mathbf{x}\sim \mathcal{D}_\text{IQA} }\left[\sum_{i=l_\text{con}}^{l-1} \log ~ \text{P}_{\theta}(\mathbf{x}_{i+1}|\mathbf{x}_1, \ldots , \mathbf{x}_i) \right],
\end{equation}
where $\mathcal{D}_\text{IQA}$ denotes the Med-IQA data and the loss is applied only to the response tokens. The two objectives are jointly optimized:
\begin{equation}
\mathcal{L}(\theta) = \mathcal{L}_{\text{IR}}(\theta) + \lambda \mathcal{L}_{\text{IQA}}(\theta),
\end{equation}
where $\lambda=0.25$. This two-stage strategy ensures that the restoration expert has already converged to a stable baseline before quality-guided conditioning is introduced, preventing the IQA loss from destabilizing early restoration training.

%% file: sections/experiments.tex
\section{Experiments}

\begin{table}[!t]
    \centering
    \caption{Restoration performance across three modalities. Best results are in \textbf{bold}.}
    \label{tab:restoration}
    \resizebox{\textwidth}{!}{%
    \begin{tabular}{l cc cc cc}
    \toprule
    & \multicolumn{2}{c}{\textbf{CT Denoising}} & \multicolumn{2}{c}{\textbf{MRI Restoration}} & \multicolumn{2}{c}{\textbf{PET Denoising}} \\
    \cmidrule(lr){2-3}\cmidrule(lr){4-5}\cmidrule(lr){6-7}
    \textbf{Method} & PSNR$\uparrow$ & SSIM$\uparrow$ & PSNR$\uparrow$ & SSIM$\uparrow$ & PSNR$\uparrow$ & SSIM$\uparrow$ \\
    \midrule
    UniMedVL~\cite{ning2025unimedvl}              & 29.82$\pm$5.12 & 0.861$\pm$0.074 & 25.93$\pm$3.10 & 0.831$\pm$0.058 & 31.65$\pm$7.05 & 0.932$\pm$0.039 \\
    Pix2Pix~\cite{isola2017pix2pix}           & 32.23$\pm$4.52 & 0.896$\pm$0.061 & 27.85$\pm$2.83 & 0.863$\pm$0.050 & 34.02$\pm$6.57 & 0.953$\pm$0.032 \\
    Restore-RWKV~\cite{yang2025restorerwkv} & 33.61$\pm$4.20 & 0.918$\pm$0.055 & 28.96$\pm$2.68 & 0.884$\pm$0.046 & 34.73$\pm$6.31 & 0.959$\pm$0.030 \\
    AMIR~\cite{amir_medical2024} & 33.70$\pm$3.76 & 0.921$\pm$0.063 & 28.62$\pm$2.71 & 0.878$\pm$0.044 & 35.08$\pm$6.24 & 0.963$\pm$0.029 \\
    MPerceiver~\cite{ai2024multimodal} & 33.85$\pm$4.03 & 0.916$\pm$0.058 & 28.47$\pm$2.75 & 0.881$\pm$0.045 & 35.21$\pm$6.18 & 0.960$\pm$0.031 \\
    MedQ-UNI (Ours)  & \textbf{36.42$\pm$4.10} & \textbf{0.935$\pm$0.050} & \textbf{29.92$\pm$2.57} & \textbf{0.905$\pm$0.042} & \textbf{36.44$\pm$6.46} & \textbf{0.974$\pm$0.028} \\
    \bottomrule
    \end{tabular}%
    }
    \end{table}
    
    \begin{figure}[t]
    \centering
    \includegraphics[width=\textwidth]{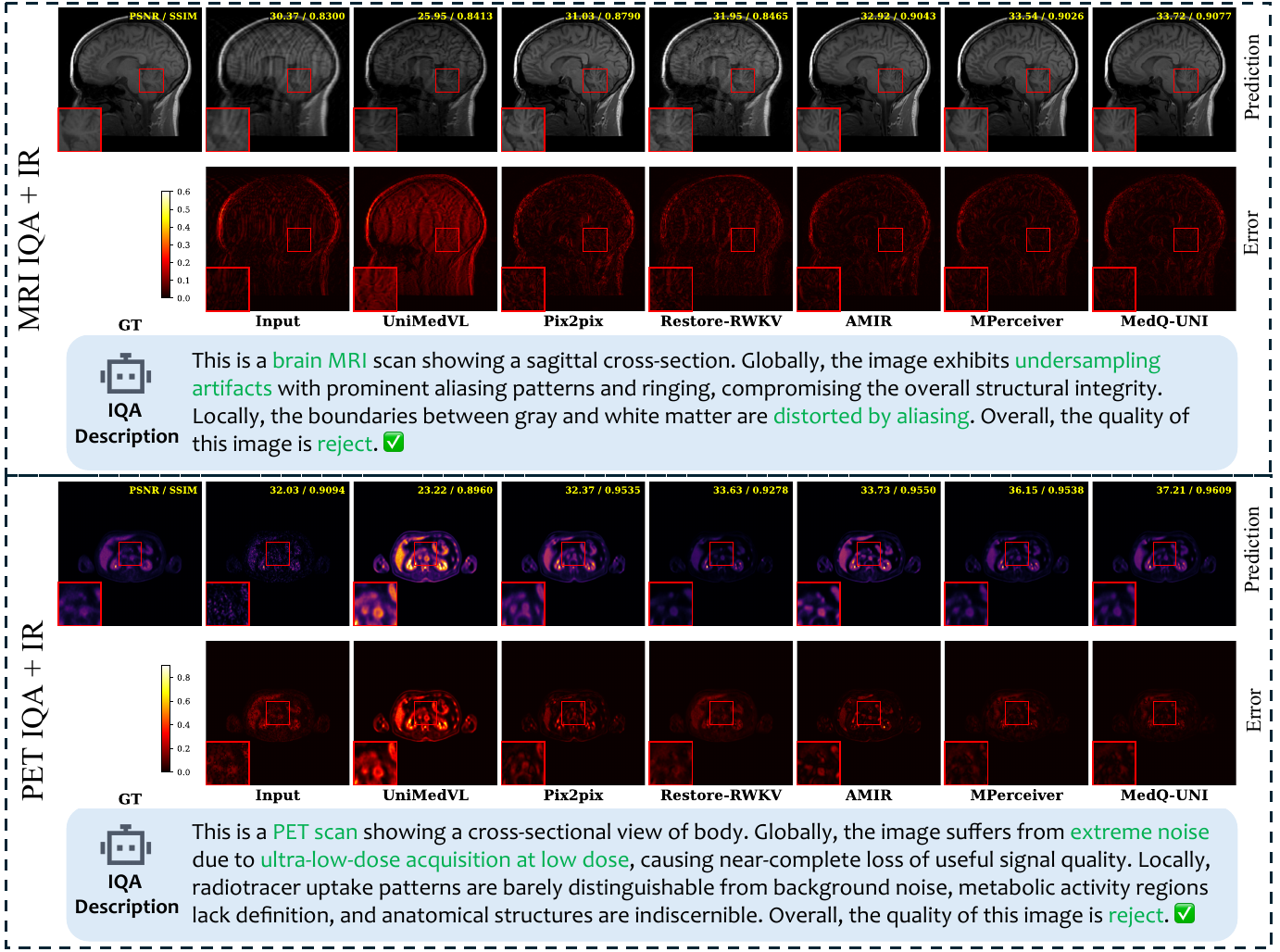}
    \caption{Qualitative comparison on MRI reconstruction (top) and PET denoising (bottom).}
    \label{fig:case}
    \vspace{-1em}
    \end{figure}

\subsection{Experimental Setup}

\noindent \textbf{Datasets.}
We use the dataset constructed in Sec.~\ref{ssec:dataset-construction}, comprising approximately 50K training pairs and an evaluation benchmark with 2K samples across three modalities (CT, MRI, PET) and five restoration tasks. 
Degradations are simulated using the TorchIO library\footnote{\url{https://docs.torchio.org/}}, including noise and motion artifacts. 
For MRI reconstruction, k-space measurements are  undersampled with $4\times$ Gaussian 1D mask.

\noindent \textbf{Evaluation metrics.}
For restoration quality we report PSNR (dB) and SSIM. For quality description evaluation we follow the MedQbench~\cite{liu2025medq} protocol and report four GPT-4o-judged dimensions: Completeness, Preciseness, Consistency, and Quality Accuracy, each scored 0--2, plus Overall as their sum.

\noindent \textbf{Implementation details.}
MedQ-UNI adopts UniMedVL~\cite{ning2025unimedvl} as our base model. We freeze the text tokenizer and VAE encoder during training and fine-tune the dual-expert modules and shared attention layers on our multi-task dataset. All experiments are conducted on 8$\times$ NVIDIA A100 GPUs. We use the AdamW optimizer with a cosine schedule from 2$\times$10$^{-5}$ initial learning rate. Stage~1 and Stage~2 are each trained for 10K steps. During inference, the rectified flow sampler uses 50 denoising steps.

\subsection{Comparison with State-of-the-Art Methods}

\subsubsection{Restoration Evaluation}

We compare against the base model UniMedVL~\cite{ning2025unimedvl}, general-purpose restoration model Pix2Pix~\cite{isola2017pix2pix}, as well as unified all-in-one methods Restore-RWKV~\cite{yang2025restorerwkv}, AMIR~\cite{amir_medical2024}, and MPerceiver~\cite{ai2024multimodal} (diffusion-based). Tab.~\ref{tab:restoration} compares MedQ-UNI against task-specific and unified baselines. MedQ-UNI achieves the highest PSNR and SSIM on every Med-IR task, with particularly notable gains on CT denoising (+2.57~dB over the best baseline) and PET denoising (+1.23~dB). On MRI restoration, MedQ-UNI also leads by a clear margin, demonstrating consistent improvements across diverse degradation types and imaging modalities. Fig.~\ref{fig:case} presents representative qualitative results on MRI and PET. By generating accurate IQA descriptions that precisely characterize degradation type and severity, MedQ-UNI adaptively conditions the restoration process, producing restorations visually closer to the ground truth with lower residual errors, particularly in fine anatomical structures. 

\subsubsection{Med-IQA Description Evaluation}

\begin{table}[t]
    \centering
    \small
    \caption{Med-IQA description evaluation.}
    \label{tab:iqa}
    \setlength{\tabcolsep}{3pt}
    \resizebox{\textwidth}{!}{%
    \begin{tabular}{lccccc|lccccc}
    \toprule
    \textbf{Method} & \textbf{Comp.} & \textbf{Prec.} & \textbf{Cons.} & \textbf{Qual.} & \textbf{Overall}
    &
    \textbf{Method} & \textbf{Comp.} & \textbf{Prec.} & \textbf{Cons.} & \textbf{Qual.} & \textbf{Overall} \\
    \midrule
    
    BiMediX2-8B~\cite{mullappilly2024bimedix2}   
    & 0.458 & 0.434 & 0.229 & 0.740 & 1.861 &
    InternVL3-38B~\cite{chen2024internvl} 
    & 1.174 & 0.909 & 1.489 & 1.456 & 5.028 \\
    
    MedGemma-27B~\cite{sellergren2025medgemma}  
    & 0.904 & 0.519 & 1.289 & 1.395 & 4.107 &
    InternVL3-8B~\cite{chen2024internvl}  
    & 1.130 & 0.968 & 1.487 & 1.456 & 5.041 \\
    
    Lingshu-32B~\cite{xu2025lingshu}   
    & 0.760 & 0.769 & 1.497 & 1.170 & 4.196 &
    Grok-4~\cite{xai2025grok4}        
    & 1.196 & 0.933 & 1.470 & 1.535 & 5.134 \\
    
    Qwen2.5-VL-7B~\cite{wang2024qwen2} 
    & 0.871 & 0.739 & 1.485 & 1.246 & 4.341 &
    Gemini-2.5-Pro~\cite{comanici2025gemini} 
    & 1.069 & 0.982 & 1.378 & 1.725 & 5.154 \\
    
    Claude-4-Sonnet~\cite{anthropic2025claude4systemcard} 
    & 0.904 & 0.698 & 1.451 & 1.521 & 4.574 &
    Qwen2.5-VL-32B~\cite{wang2024qwen2} 
    & 1.276 & 1.023 & 1.500 & 1.426 & 5.225 \\
    
    Mistral-Medium-3~\cite{jiang2023mistral7b} 
    & 1.124 & 0.804 & 1.278 & 1.480 & 4.686 &
    GPT-4o~\cite{achiam2023gpt}        
    & 1.229 & 1.132 & 1.492 & 1.555 & 5.408 \\
    
    Qwen2.5-VL-72B~\cite{wang2024qwen2} 
    & 1.102 & 0.948 & 1.494 & 1.460 & 5.004 &
    \textbf{MedQ-UNI} 
    & \textbf{1.284} & \textbf{1.203} & \textbf{1.507} & \textbf{1.836} & \textbf{5.830} \\
    
    \bottomrule
    \end{tabular}}
    \vspace{-6pt}
    \end{table}

Tab.~\ref{tab:iqa} compares MedQ-UNI against medical-specific and general-purpose vision-language models. MedQ-UNI achieves the highest scores on all dimensions, outperforming even the strongest closed-source model. This confirms that domain-specific training on medical quality descriptions yields substantially more accurate assessments than models that lack medical quality priors.

\subsection{Ablation Study}

\begin{table}[t]
\centering
\caption{Ablation study. All variants share the same architecture and training data unless otherwise noted.}
\label{tab:ablation}
\resizebox{\textwidth}{!}{%
\begin{tabular}{l cc cc cc}
\toprule
& \multicolumn{2}{c}{\textbf{CT Denoising}} & \multicolumn{2}{c}{\textbf{MRI Restoration}} & \multicolumn{2}{c}{\textbf{PET Denoising}} \\
\cmidrule(lr){2-3}\cmidrule(lr){4-5}\cmidrule(lr){6-7}
\textbf{Variant} & PSNR$\uparrow$ & SSIM$\uparrow$ & PSNR$\uparrow$ & SSIM$\uparrow$ & PSNR$\uparrow$ & SSIM$\uparrow$ \\
\midrule
\multicolumn{7}{l}{\textit{Loss design}} \\
w/o pixel + SSIM loss       & 36.04$\pm$4.18 & 0.919$\pm$0.055 & 29.67$\pm$2.63 & 0.889$\pm$0.046 & 36.12$\pm$6.44 & 0.964$\pm$0.031 \\
\midrule
\multicolumn{7}{l}{\textit{IQA conditioning}} \\
w/o IQA description         & 35.47$\pm$4.22 & 0.921$\pm$0.054 & 29.18$\pm$2.66 & 0.890$\pm$0.045 & 35.62$\pm$6.50 & 0.963$\pm$0.030 \\
\midrule
\multicolumn{7}{l}{\textit{Quality description quality}} \\
Noisy pseudo-label desc.    & 36.10$\pm$4.16 & 0.929$\pm$0.052 & 29.63$\pm$2.60 & 0.898$\pm$0.043 & 36.18$\pm$6.45 & 0.970$\pm$0.029 \\
Expert-refined desc. (Ours) & \textbf{36.42$\pm$4.10} & \textbf{0.935$\pm$0.050} & \textbf{29.92$\pm$2.57} & \textbf{0.905$\pm$0.042} & \textbf{36.44$\pm$6.46} & \textbf{0.974$\pm$0.028} \\
\midrule
\multicolumn{7}{l}{\textit{Training strategy}} \\
End-to-end joint training   & 35.83$\pm$4.30 & 0.922$\pm$0.056 & 29.35$\pm$2.70 & 0.891$\pm$0.047 & 35.90$\pm$6.52 & 0.965$\pm$0.031 \\
Two-stage (Ours)            & \textbf{36.42$\pm$4.10} & \textbf{0.935$\pm$0.050} & \textbf{29.92$\pm$2.57} & \textbf{0.905$\pm$0.042} & \textbf{36.44$\pm$6.46} & \textbf{0.974$\pm$0.028} \\
\bottomrule
\end{tabular}%
}
\vspace{-5pt}
\end{table}

To validate the contribution of each design choice in MedQ-UNI, we conduct ablation experiments across four dimensions (Tab.~\ref{tab:ablation}). \textit{Loss design:} Removing pixel and SSIM losses yields the largest SSIM drops, confirming their direct role in structural preservation. \textit{IQA conditioning:} Removing the IQA description produces substantial drops across all modalities, confirming that the explicit textual quality description provides essential degradation-aware guidance that significantly benefits restoration beyond the implicit quality information transferred through shared attention. \textit{Quality description quality:} We compare two levels of description quality: (i) noisy pseudo-label descriptions generated by Qwen3-VL-235B without expert refinement, and (ii) expert-refined descriptions used in the full pipeline. Performance improves with description quality, demonstrating that richer and more accurate quality descriptions translate directly into better restoration. \textit{Training strategy:} End-to-end joint training from scratch degrades restoration performance compared to our two-stage strategy, confirming that pre-training the restoration expert before introducing the IQA loss prevents early-stage optimization instability.

%% file: sections/conclusion.tex
\section{Conclusion}

We presented the \emph{assess-then-restore} paradigm and its instantiation, MedQ-UNI, a multimodal autoregressive vision-language model that unifies medical image quality assessment and restoration within a single dual-expert framework. Supported by a 50K-sample dataset with structured quality descriptions across three modalities and five tasks, MedQ-UNI achieves state-of-the-art restoration while producing quality descriptions that surpass leading vision-language models. Ablation studies confirm that explicit quality conditioning and pixel-level losses both contribute to these gains, validating that coupling quality understanding with restoration yields both stronger performance and interpretable outputs. A key strength of MedQ-UNI is its generalizability: by autonomously assessing degradation type and severity from the input image itself, the model can perform targeted restoration across arbitrary modalities and degradation types within a single network, without requiring task-specific models or manual degradation specification. Future work will extend MedQ-UNI to additional modalities and investigate interactive quality feedback mechanisms.